\pdfoutput=1

\documentclass[11pt]{article}

\usepackage{acl}

\usepackage{times}
\usepackage{latexsym}

\usepackage{booktabs}

\usepackage{paralist}

\usepackage[T1]{fontenc}

\usepackage[utf8]{inputenc}

\usepackage{microtype}

\usepackage{amsmath}
\usepackage{pifont}
\newcommand{\cmark}{\ding{51}}%
\newcommand{\xmark}{\ding{55}}%

\usepackage{inconsolata}

\usepackage{xcolor}
\definecolor{apricot}{HTML}{FFD982}
\definecolor{yellowgreen}{HTML}{98CC70}
\definecolor{thistle}{HTML}{D883B7}
\usepackage{csquotes}

\usepackage{graphicx}

\title{\textsc{CHIRON}: Rich Character Representations in Long-Form Narratives}

\author{Alexander Gurung \\
  School of Informatics \\
  University of Edinburgh \\
  Edinburgh, UK \\
  \texttt{a.gurung-1@sms.ed.ac.uk} \\\And
  Mirella Lapata \\
  School of Informatics \\
  University of Edinburgh \\
  Edinburgh, UK \\
  \texttt{mlap@inf.ed.ac.uk} \\}

\begin{document}
\maketitle
\begin{abstract}

Characters are integral to long-form narratives, but are poorly
understood by existing story analysis and generation systems. While
prior work has simplified characters via graph-based methods and brief
character descriptions, we aim to better tackle the problem of
representing complex characters by taking inspiration from advice
given to professional writers. We propose \textsc{CHIRON}\footnote{In
  Greek mythology, CHIRON was a highly educated centaur known for his wisdom, and as a mentor of
  many Greek heroes, including Achilles.}, a new `character-sheet'
based representation that organizes and filters textual information
about characters.  We construct \textsc{CHIRON} sheets in two steps: a
Generation Module that prompts an LLM for character information via
question-answering and a Validation Module that uses automated
reasoning and a domain-specific entailment model to eliminate false
facts about a character.  We validate \textsc{CHIRON} via the
downstream task of masked-character prediction, where our experiments
show \textsc{CHIRON} is better and more flexible than comparable
summary-based baselines. We also show that metrics derived from
\textsc{CHIRON} can be used to automatically infer
character-centricity in stories, and that these metrics align with
human judgments.\footnote{Our data and code are available at

\href{https://github.com/Alex-Gurung/CHIRON}{https://github.com/Alex-Gurung/CHIRON}
}

\end{abstract}


\section{Introduction}


\begin{figure*}[t]
    \centering
    \includegraphics[width=1\textwidth]{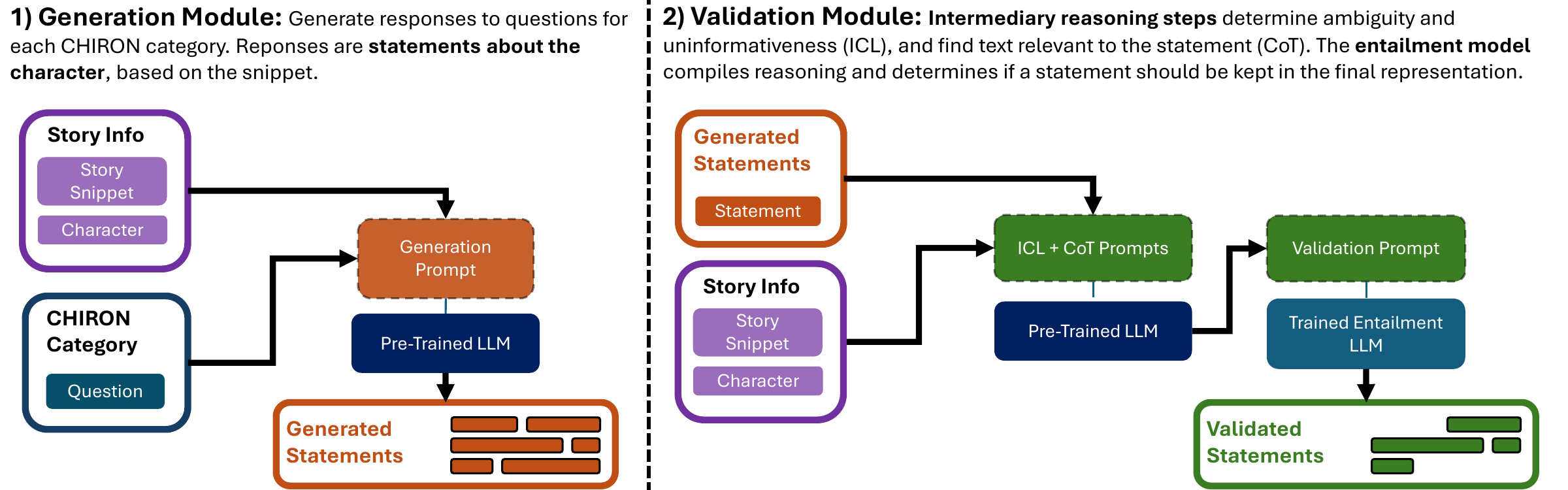}
   \vspace*{-.2cm}
    \caption{Overview of \textsc{CHIRON}'s Generation and Validation Modules. For each \textsc{CHIRON} category, our Generation Module takes a character and story snippet and uses a pretrained LLM to generate statements about the character. 
    Our Validation Module then passes these statements through reasoning steps and a trained entailment classifier to determine if they are true and useful. 
    We find this methodology produces more accurate and useful character-representations for downstream tasks and analysis.
    We use Mistral 7B-Instruct v0.2 unless otherwise stated. 
    } 
    \label{fig:front_page2}
\end{figure*}

\begin{displayquote}
``The three things that matter most in a story are characters,
  characters and characters.'' -- Bob Gale\footnote{American
  screenwriter, producer, and film director best known for co-writing
  the "Back to the Future" trilogy.}
\end{displayquote}

Advice to professional writers emphasizes the importance of
well-developed and interesting characters in stories, identifying them
as the key to engaging and believable narratives
\citep{kyle_page-turner_2016, jarvis_crafting_2014,
  weiland_creating_2016,
  the_editors_of_writers_digest_books_creating_2014}.

Computational techniques aiming to analyze and generate literary texts
have also emphasized the importance of characters in storytelling by
tracking their relationships and emotional trajectories
\cite{chaturvedi2017unsupervised,iyyer-etal-2016-feuding,elsner-2012-character},
identifying enemies and allies \cite{nalisnick-baird-2013-character},
villains or heroes
\cite{bamman-etal-2013-learning,bamman-etal-2014-bayesian}, and
representing characters via attributes \cite{Zhang2019GeneratingCD},
personality types \cite{sang2022mbti}, or 
high-level paragraph-style descriptions
\citep{brahman-etal-2021-characters-tell, chen_persona_2024,
  li_zero-shot_2024, xu_character_2024, yu_few-shot_2022,
  yang_doc_2023, yang_re3_2022}.




In this work we propose a new `character-sheet' based representation,
\textsc{CHIRON}, standing for \textbf{CH}aracter-\textbf{I}nformed
\textbf{R}epresentations \textbf{O}ver
\textbf{N}arratives. \textsc{CHIRON} consists of structured
descriptions of characters along several dimensions, encapsulating
character nuance and complexity, while still being interpretable and
useful for downstream tasks and analysis.  It is composed of two
modules: a Generation Module that generates long-form answers to
questions based on a snippet of story, and a Validation Module that
verifies the utility and truthfulness of these answers.

Drawing inspiration from how writers create their own characters
\citep{jarvis_crafting_2014,weiland_creating_2016,the_editors_of_writers_digest_books_creating_2014},
our character sheets have four broad categories: \textbf{Dialogue}
(how a character speaks), \textbf{Physical/Personality} (how a
character looks/acts), \textbf{Knowledge} (what a character knows),
and \textbf{Goals} (what a character wants to accomplish/has
accomplished thus far).  We fill these categories (see
Table~\ref{tab:character_sheet_categories}) in the Generation Module
with information about a given character by prompting out-of-the-box
language models with a snippet from a story and questions relevant to
each category.  Despite using a wide variety of models, we find that
the generations at this stage are often unfaithful to the story. We
develop a Verification Module that uses automated reasoning in
combination with a high-precision entailment model to ensure our
character-representations are accurate and useful.

We build this work on STORIUM \citep{akoury_storium_2020}, a dataset
of collaboratively written stories where contributors write entries
from the perspective of a specific character. This gives us a perfect
test-bed to explore character representations, as we know which
character will be heavily featured in each story entry.
%
%
%
We validate \textsc{CHIRON} in two ways. We assess its usefulness for
the downstream task of masked-character prediction, generating
character sheets up-until a given story snippet and then predicting
the character-IDs from the masked snippet. We find an $11.6\%$
performance improvement over summary-based baselines, demonstrating
that our structured format would be more useful than existing methods
for character-centric tasks.  We also develop automated metrics based
on our character sheets as a means of understanding the use of
characters in human-written
\cite{akoury_storium_2020,chakrabarty_art_2024} and \mbox{LLM-generated}
stories \cite{yang_doc_2023}. We show that our proposed
\textit{density} metric aligns with human judgments of
character-centricity in stories.
%
%
Our primary contributions are:

\begin{itemize}
   \setlength\itemsep{-0.2em}
  \vspace{-0.2cm}
    \item We present \textsc{CHIRON}, a novel `character-sheet'
      structure, for representing characters in long-form narratives;
    \item We develop a Validation Module to automatically determine if
      a given statement about a character is true based on the
      snippet, and use it to produce more accurate character sheets
      over the course of a story;
    \item We validate \textsc{CHIRON} via the  task of
      masked-character prediction, showing improved performance over competitive baselines;
    \item We use automated metrics derived from \textsc{CHIRON} to better understand characters in human-written and LLM-generated stories.
\end{itemize}

\section{Related Work}

\paragraph{Modeling Stories, Narratives and Characters}

Prior work represents characters in three general ways: 1)~as discrete
values (e.g.,~sentiments, adjectives, or graph nodes)
\citep{hoque_portrayal_2023, soni_grounding_2023, yu_personality_2023,
  gorinski_movie_2015, elson_extracting_2010, ye_neural_2022,
  peng_guiding_2022} 2)~as embeddings \citep{bamman_learning_2013,
  kim_learning_2018} and 3)~as high-level paragraph-style descriptions
or summaries \citep{brahman-etal-2021-characters-tell,
  chen_persona_2024, li_zero-shot_2024, xu_character_2024,
  yu_few-shot_2022, yang_doc_2023, yang_re3_2022}. Our work is most
similar in form to this last approach, and we use generated character
summaries as a baseline in our experiments.

Much research on modeling characters comes from role-playing in
text-based games \citep{ammanabrolu_how_2021, shuster_deploying_2020,
  zhu_fireball_2023,
  rameshkumar_storytelling_2020,louis_deep_2018}. In particular,
\citet{ammanabrolu_how_2021} showed good role-playing dialogue and
action performance by providing models with multi-faceted motivations
and brief character summaries.
%
%
%
Early work using LLMs for story generation has explored the benefit of
character information indirectly by providing plot outlines as
guidance
\cite{rashkin_plotmachines_2020,fan_hierarchical_2018,peng_guiding_2022}.
More recent work focusing on generating longer stories gives more
weight to complex plot structure \citep{yang_doc_2023, yang_re3_2022,
  chung_talebrush_2022}. \citet{yang_doc_2023} tracks character
development while generating stories, although their representation of
characters is unstructured and difficult to validate independently.


\paragraph{Literary Understanding of Characters}

We take inspiration from existing discussion aiming to help writers
develop characters in their own narratives
\citep{jarvis_crafting_2014,weiland_creating_2016,the_editors_of_writers_digest_books_creating_2014}. While
the advice is largely meant for \textit{creating} fully fleshed out
characters, it is also useful for \textit{analysing} existing ones,
and provides insight into what ancillary notes writers use to keep
characters consistent. Two insights in particular greatly inform this
work 1)~the aspects important to different characters vary wildly and
can be extremely nuanced in their substance and 2)~despite this, many
writers still find it useful to keep organized notes on key character
details as they write. We interpret these insights to indicate the
need for un-simplified descriptions of character features, and
hypothesize that an organized collection of these descriptions would
be useful for analysing and generating stories.

\section{STORIUM Dataset}
\label{sec:Storium}

We base our work on the STORIUM dataset, a collection of 6k stories
written collaboratively online \citep{akoury_storium_2020}. We filter
these for length (avoiding stories with too little content or too long
for analysis), English language, and traditional novel writing style
(as opposed to game-playing). 
This leaves us with 529 stories, containing an average of 5.5k words
in 42 entries. Each entry is from a  specific character's perspective
(although as in traditional narratives the use of grammatical person
and narrator omniscience varies). Due to the length of many of these
entries, we further split them into snippets, along paragraph
lines. We select the top three most frequently occurring characters,
giving us 1,403~character-story combinations and
9,021~snippets. Further details are presented in
\autoref{sec:appendix_storiumfiltering}.

While our \textsc{CHIRON} representation can be applied to any
story text, we primarily rely on STORIUM for development and
validation. As each of these story snippets is centered around the
given character, STORIUM is ideal for exploring how characters are
represented in narratives. An example STORIUM snippet and
character sheet based on it is shown in
\autoref{tab:character_sheet_categories}.

\begin{table*}[t]
  \centering
  \small 
  \begin{tabular}{@{}p{15.3cm}@{}} \toprule
\multicolumn{1}{@{}c@{}}{\textbf{Story Snippet}} \\ \midrule
 Nadia watched both of the men silently, pulling her tattered,
  too-big sweatshirt tighter around herself. She hadn't spoken since
  the plane had crashed, and she didn't want to, but it looked like
  the men were about to devolve into an argument about whether or not
  to go into the cave, and she actually had input.  She cleared her
  throat and, when she had the attention of both Jacob and Jayson, she
  pointed to the sky. Their heads all turned.  They could barely see
  them through the trees, but the grey storm clouds were unmistakable
  all the same.  Nadia tried to speak but had to clear her throat once
  more and swallow before she was able to do so, not having used her
  voice for so long. "I don't know about you, but I'd really rather
  not be out in the forest during a storm. I say we check out the
  cave. If we're careful about it, we won't run into any trouble." \\ \bottomrule
  \end{tabular}
  \begin{tabular}{@{}p{1.6cm}p{6.6cm}p{6cm}@{}}
    \textbf{Dialogue} &  What, if anything have we learned about how this
    character speaks from this snippet? & She has gone without using her voice for an extended
    period.\\ \midrule
    \textbf{Physical/ Personality} & What, if any, physical descriptions of this
    character are in this snippet? & She pulls her tattered, too-big sweatshirt tighter around herself.\\
     & What, if any, descriptions of this
    character's personality are in this snippet? & Nadia is quiet. \\\midrule
 \textbf{Knowledge} & 
What, if any, factual information is given about this character in
this snippet? &  Nadia is a survivor of a plane crash. \\
& What, if any, information has this character learned in this
snippet? &  She is aware of the approaching storm. \\ \midrule
\textbf{Plot} & 
 What, if any, goals does this character gain in this snippet that
 they wish to accomplish in the future? & Nadia's goal in this snippet
 is to seek shelter from the approaching storm. \\ 
 & What, if any, goals does this character complete in this
 snippet?  &  Nadia suggests seeking shelter in the cave during an approaching storm. \\
&  How, if at all, does this character's internal motivations change
    in this snippet? & Nadia, who had been silent since the plane
    crash, feels the need to speak up. \\ \bottomrule
  \end{tabular}
\caption{\label{tab:character_sheet_categories} Sample character sheet
  for Nadia. Answers to questions were provided
  by Mistral 7B-Instruct v0.2.}
\end{table*}

\section{\textsc{CHIRON}: Character Sheets}

While prior work has used detailed plot outlines and knowledge graphs
\citep{peng_guiding_2022,yang_doc_2023}, developing rich character
representations in a manner similar to writers is an unexplored area
of research. Such representations should accurately describe all
attributes of a given character that may be important for the reader
to know, and should be consistent (or change logically) across the
story. Some attributes (e.g.,~height, name) are well suited for
knowledge graphs and other database-inspired representations, but many
of the more subtle facts about characters are too complex to store in
this manner (e.g.,~they are self-conscious about their
clothes). 

Inspired by advice for writers
\citep{jarvis_crafting_2014,weiland_creating_2016,the_editors_of_writers_digest_books_creating_2014},
we propose instead to represent characters in a long-form
character-sheet structure we call \textsc{CHIRON}. We extrapolated themes and condensed lists of specific questions (e.g. ``Is this character intelligent, and in what way?'') from these resources to create a structure more broadly applicable to any story.
As seen in \autoref{tab:character_sheet_categories}, \textsc{CHIRON} is organized as four broad categories, each composed of questions-and-answer pairs:






\begin{asparadesc}
\item[\textbf{Dialogue:}] how the character speaks; this could include
accents, catch-phrases, whether they prefer flowery or direct prose,
etc.

\item[\textbf{Physical/Personality:}] descriptions of the character's
 appearance and personality; this could include height, scars,
courage, condescension, etc.

\item[\textbf{Knowledge:}] other factual details about this character, as well as knowledge this character has learned about the world; this could
include family members, a tragic backstory, where the gold is hidden,
etc.

\item[\textbf{Goals:}] goals this character has gained or completed, and
changes in their internal motivation; this could include finding the
treasure, gaining a quest, wanting revenge, etc.
\end{asparadesc}

We operationalize \textsc{CHIRON}'s character-representation task as
two steps: a Generation Module and a Validation Module. (i) The
generation step takes in a character and snippet of story, and uses an
out-of-the-box LLM to answer a series of open-ended questions about
the character, and organizes these question-answer pairs into
categories. (ii) The validation step then splits the resulting answers
into sentences and filters out incorrect sentences using our
intermediary reasoning steps and domain-specific entailment
model. \autoref{fig:front_page2} shows an overview of this two-step
process and \autoref{tab:character_sheet_categories} shows an example
character sheet (one sentence taken per question).

This method allows us to leverage the high performance of out-of-the-box LLMs on instruction following tasks, while using our entailment model to prevent hallucinations corrupting the resulting representation.
For tasks that require a character-representation
across multiple snippets, we simply concatenate these filtered answers
by question and remove duplicates via pairwise cosine similarity
between TF-IDF vectors (thresholded at~$0.9$).
We also incorporate a simplification step prior to the Validation Module that splits sentences into individual claims; further explanation is in \autoref{sec:appendix_prompt} and \autoref{tab:simplifying_generating_prompt_template}.


\section{Validation Module}
\label{sec:validation-module}

Given a story snippet, character, and statement (about said
character), our Validation Module predicts whether the statement
entails completely from the snippet. As this task is extremely
domain-specific, we construct this module from two novel parts:
automated reasoning and a trained entailment model.  In this section,
we describe our methodology for collecting data, setting up automated
reasoning, and training our domain-specific entailment model. 

\subsection{Dataset Creation and Annotation}

\paragraph{Character Sheet Generation} We generate character sheets by
showing a pretrained out-of-the-box LLM a snippet of story and the
associated character, and asking questions for each category. Full
prompt examples are shown in \autoref{sec:appendix_prompt}. To
diversify our responses, we generate character sheets using four
models of varying sizes: Mistral 7B Instruct v0.2
\citep{jiang_mistral_2023}, Llama~2 Chat 13B
\citep{touvron_llama_2023}, Tulu~2 DPO 70B \citep{ivison_camels_2023},
and GPT-4 (version gpt-4-0613) \citep{achiam_gpt-4_2023}. We generate
250 character sheets from each model, each representing a unique
story-character combination. Taking the question-response pairs that
compose character sheets, we split responses into sentences to give us
sentence-snippet-character tuples. We randomly sample 1,000 tuples
(250 from each model) for annotation.

\paragraph{Entailment Annotations}
Human annotators were asked to judge whether a statement about a
character entails from a story snippet using a 1--5~scale to account
for ambiguity and multiple interpretations of the same text. A
statement labeled with: 1~is entirely unsupported by the snippet; 2~is
largely contradicted by the snippet; 3~is ambiguous in its
relationship with the snippet; 4~is likely true based on the snippet,
or 5~is entirely supported by the snippet.
Detailed examples are included in \autoref{sec:appendix_annotation},
\autoref{tab:example_annotations}.

Annotators were recruited through Prolific; they were first shown
instructions and a series of examples with justifications for the
labels, before annotating five snippet-statement pairs on their
own. In addition to entailment (called `consistency' for
comprehension), crowdworkers were asked to annotate statement
coherence and to provide brief justifications for their answers, which
we found improved attention to detail. Screenshots of the interface
and more details are in \autoref{sec:appendix_annotation}. Our data
was annotated by a group of~7 annotators who passed various screening
tests and spent a considerable amount of time on the task (over 21
minutes on average).  Inter-annotator agreement was high, achieving a
Krippendorff Alpha of $0.679$.
 

\autoref{tab:dataset_dist} shows the label distribution in our
annotated dataset, we find that $52\%$ of sentences were entirely
correct and entailed from the snippet, and an additional
$15.4\%+52\%=67.4\%$ were largely correct (a label of 4 generally
indicates a minor potentially ambiguous or unclear phrase amongst an
otherwise verifiably correct statement). The remaining $32.6\%$ of
sentences were unverifiable or false, although it should be noted that
this includes statements that make no claims. Further details and
examples are presented in \autoref{sec:appendix_examples}. 

\begin{table}[t]
\centering
\small
\begin{tabular}{p{0.13\textwidth} | p{0.04\textwidth} p{0.04\textwidth} p{0.04\textwidth} p{0.04\textwidth} p{0.04\textwidth}}
\toprule
\textbf{Label: } & \multicolumn{1}{c}{\textbf{1}} & \multicolumn{1}{c}{\textbf{2}} & \multicolumn{1}{c}{\textbf{3}} & \multicolumn{1}{c}{\textbf{4}} & \multicolumn{1}{c}{\textbf{5}} \\
\midrule
Mistral 7B & 10.8 & 2.8 & \hspace*{.13cm}9.6 & 15.2 & 61.6 \\
Llama-2 13B & 24.4 & 3.2 & 13.6 & 17.2 & 41.6 \\
Tulu-2 70B & 10.4 & 2.4 & 12.4 & 16.8 & 58.0 \\
GPT-4 & 24.0 & 1.6 & 14.4 & 13.2 & 46.8 \\
\midrule
Combined & 17.4 & 2.5 & 12.7 & 15.4 & 52.0 \\
\bottomrule

\end{tabular}
\caption{\label{tab:dataset_dist} Distribution of labels (in \%) in
  annotated dataset (1,000 examples, 250 per model). Despite being the
  smallest model, Mistral 7B-Instruct seems to perform the best,
  although prompts were not optimized for each model, so further work
  could find different results. Prompts were primarily tested using
  Mistral and Llama before being used uniformly across all models. }
\end{table}


\subsection{Statement Validation}
\label{sec:statement-validation}


We split our dataset 80-10-10 for train-validation-test, and use
Mistral 7B Instruct v0.2 \citep{jiang_mistral_2023} as our base model,
given its better entailment performance (see
Table~\ref{tab:dataset_dist}). Although overall accuracy is important,
we pay particular attention to precision as we want to ensure all
statements included in the final character sheet are correct.

\paragraph{Automated Reasoning}

Prior work has shown In-Context-Learning (ICL) and Chain-of-Thought
(CoT) reasoning can improve zero-shot classification performance
\citep{brown_language_2020,wei_chain_2022}. To this end, we generate
intermediate reasoning steps using Mistral 7B Instruct v0.2
\citep{jiang_mistral_2023} that help our classifier accurately
identify entailment by plainly decomposing the reasoning steps. We
split reasoning into three steps: ambiguity (is this statement too
ambiguous to verify), informative (does this statement provide any
information about the character), and a chain-of-thought step that
prompts the model to retrieve relevant pieces of the story and then to
compare and contrast with the statement. More details are given in
\autoref{sec:appendix_automatedreasoning}.

\paragraph{Domain-Specific Entailment Model}

\begin{table}[t]
\small
\centering
\begin{tabular}{lccc}
\toprule
\textbf{Name} & \multicolumn{1}{c}{\textbf{Precision}} & \multicolumn{1}{c}{\textbf{Recall}} & \multicolumn{1}{c}{\textbf{Accuracy}} \\
\midrule
Zero-Shot All & 0.753 & \textbf{0.986} & 76\% \\
Zero-Shot ICL & 0.759 & 0.887 & 72\% \\
Zero-Shot CoT & 0.753 & \textbf{0.986} & 76\% \\
Zero-Shot None & 0.835 & 0.930 & \textbf{82\%} \\
Trained All & 0.930 & 0.746 & 78\% \\
Trained ICL & \textbf{0.952} & 0.563 & 67\% \\
Trained CoT & 0.864 & 0.803 & 77\% \\
Trained None & 0.895 & 0.718 & 74\% \\
\bottomrule
\end{tabular}
\caption{\label{tab:classifier_performance}
Comparing baseline (zero-shot) performance and trained performance by shown reasoning steps. `All' refers to showing both ICL and CoT reasoning steps to the model, ICL and CoT refers to just showing their respective reasoning steps, and `None' refers to not showing any reasoning steps. We prioritize precision but want to maintain a good balance between precision and recall, so choose the `Trained All' setting moving forward. All models are based on Mistral 7B Instruct v0.2 \citep{jiang_mistral_2023}.}
\end{table}

We combine the snippet, reasoning steps, character, and statement in
the context and prompt a Mistral 7B model for an entailment label
(1--5). To best take advantage of the language model's existing
abilities we fine-tune with the causal language modeling objective (on
just this labeling step) and LoRA ($r=16,\alpha=32$)
\citep{hu_lora_2021}. During training we take the best-performing
model by precision (boolean where label $ \geq 4$) on the validation
set. We find slightly improved performance by thresholding our model's
predictions at $=5$, further details and comparisons are presented in
\autoref{sec:appendix_classifier_details}.

\autoref{tab:classifier_performance} compares our trained model
(Trained All) against ablated variants (ICL, CoT, None) and zero-shot
models.  As can be seen, Trained~All improves on baseline zero-shot
precision (without reasoning) by~$0.095$, achieving a precision
of~$0.930$. Removing our CoT step can further improve
precision, but at too great a cost to recall and accuracy.  We find
the performance of this Validation Module promising as it shows that
even in this rhetorically complex domain where models of varying sizes
struggle to extract consistent character details, the analytical task
of determining entailment is possible simply by fine-tuning a
7B-parameter model. As we use this Validation Module to filter out
incorrect statements, the high precision also lends credence to the
resulting character sheets without needing to approach the more
difficult task of improving character-statement generations.
%
%
An example of a full character sheet (across an entire story) before
and after our Validation Module is shown in
\autoref{sec:appendix_full_character_sheet_examples}
(\autoref{tab:full_character_sheet_p1},
\autoref{tab:full_character_sheet_p2}).


\section{Masked Character Prediction}
\label{sec:mask-char-pred}

In this section we validate \textsc{CHIRON} on a masked character
prediction task \citep{holgate-erk-2021-politeness, brahman_let_2021}
by showing it serves as a better character-representation than a
competitive baseline. We define the task as follows: given an unseen
STORIUM story, we get the three most common characters by scene
appearance and identify story snippets that contain all three
characters. We create character sheets 
for each story up-to this snippet. We mask out character names from
the story snippet and, given this masked story snippet and generated
character-information, prompt Mistral 7B Instruct v0.2
\citep{jiang_mistral_2023} to match each character to its
mask-ID. Filtering for finished, previously unseen stories and scenes with their three most common characters gives us a dataset of 49 entries with 3 characters each,
or 147 data points. \autoref{tab:masked_storium_example} shows an
example.

\begin{table}[t]
\centering
\footnotesize
\begin{tabular}{p{0.95\linewidth}}
\hline
\colorbox{apricot}{[CHAR 0]} glanced over when \colorbox{thistle}{[CHAR 2]} mentioned \colorbox{yellowgreen}{[CHAR 1]}. "Heh, well she better get going if she doesn't want to be left behind." \colorbox{apricot}{[CHAR 0]} caught his breath as he leaned down, taking a moment, making sure he was hidden from sight. He saw \colorbox{yellowgreen}{[CHAR 1]} lean down and hold Ian for some time, before getting up and calmly walking away. "Wha'she doing there ya think? Speaking with the dead? Ah, whatever." \\

\colorbox{thistle}{[CHAR 2]} piped up about his uneasiness in this world. \colorbox{apricot}{[CHAR 0]} scoffed at him. "What does it matter? It's just a tree. I haven't noticed anything different. I'm sure you're just shaken from seeing a guy murdered next to you," \colorbox{apricot}{[CHAR 0]} chuckled as he smacked \colorbox{thistle}{[CHAR 2]} on the back. "Now, where to next, mr. navigator?" \\
\hline
\end{tabular}
\caption{\label{tab:masked_storium_example} Masked snippet from unseen
  STORIUM story. The task is as follows: given a character name and
  representation, determine which character-ID (0, 1, or~2)
  corresponds to that character in the snippet.}
\end{table}

\begin{table*}[t]
\footnotesize
  \centering
\begin{tabular}{l | l}
\hline
\textbf{Summary Question}   &  \textbf{Response}        \\
\hline
{\small \begin{minipage}[t]{0.98\columnwidth}
Summarize everything we have learned about this character across these snippets. Include aspects of the character like how they speak, what they look like, their personality, their goals, etc.
\end{minipage}}      & {\small \begin{minipage}[t]{0.98\columnwidth} Nadia is a quiet and thoughtful survivor. She wears a tattered, too-big sweatshirt. She hasn't spoken much since the plane crash. She suggests checking out the cave for shelter. She's resourceful and thinks logically. She's determined to survive and keep the group safe.\end{minipage}}     \\
\hline
\end{tabular}
\caption{\label{tab:summary_baseline_example}
Sample \textit{Character-Summary} baseline for Nadia. Input story snippet was the same as in \autoref{tab:character_sheet_categories}.}
\end{table*}

\begin{table}[t]
{\small
\centering
\begin{tabular}{lcc}
\hline
\textbf{Setup} & \textbf{OnlyRole} & \textbf{Accuracy} \\
\hline
NoInformation & --- & 43.5\% \\
Character-Summary & \xmark & 44.9\% \\
\textsc{CHIRON:} Entire-Sheet & \xmark & 47.6\% \\
\textsc{CHIRON:} Agreed & \xmark & \textbf{50.3\%} \\
\hline
Character-Summary & \cmark & 42.2\% \\
\textsc{CHIRON:} Entire-Sheet & \cmark & \textbf{56.5\%} \\
\textsc{CHIRON:} Agreed & \cmark & 53.7\% \\
\hline
\textsc{CHIRON:} Agreed-Subset & \cmark & \textbf{58.5\%} \\
\hline
\end{tabular}
\caption{\label{tab:masked_prediction_results} Accuracy on
  masked-character prediction task, separated by \textit{OnlyRole}
  (whether to only include snippets from the characters perspective
  when creating a character representation).  \textsc{CHIRON}
  representations uniformly perform better than \textit{NoInformation}
  and \textit{Character-Summary} baselines; this difference becomes
  more pronounced under the \textit{OnlyRole} setting. We also report
  the best performing \textit{Agreed} setting across all possible
  subsets of categories (Knowledge+Physical). } }
\end{table}

\paragraph{Settings and Baselines}

We have two CHIRON-based settings.  \textit{EntireSheet} is the most
traditional setting, where the entire CHIRON character-sheet is shown
to the model.  In contrast, our \textit{Agreed} setting refers to
showing to the masked-character prediction model only one category at
a time (Dialogue, Physical/Personality, Knowledge, or Goals), and
aggregating the predictions by multiplying the softmax probabilities
and taking the maximum. We also briefly explore aggregating the
predictions from a subset of our categories, as some may be more
informative than others for this specific task.

We also have two baseline settings. \textit{Character-Summary} sets
the character-information to a single character description obtained
by prompting Mistral 7B Instruct v0.2 \citep{jiang_mistral_2023} with
the entire story so-far and asking for a summary of the given
character. This summary is then filtered using our entailment
classification pipeline to ensure that differences in performance can
be attributed to the difference in representation (as opposed to
differences in hallucination
rates). \autoref{tab:summary_baseline_example} presents an example
summary. \textit{No-Information} refers to showing no
character-information during the masked-character prediction task. As
pronouns and other name-based clues make the task slightly easier than
random, this baseline serves as a more accurate performance floor.

We also introduce a STORIUM-specific hyper-parameter \textit{OnlyRole}
that applies to all character representations. \textit{OnlyRole}
refers to the boolean choice of whether to only include story snippets
from the character-of-interest's perspective as input (where
\textit{true} refers to only using such snippets). Excluding
non-perspective snippets may leave out crucial information, but may
increase the density of useful information for the task. We ultimately
report performance under both settings and find \textsc{CHIRON} tends
to benefit from \textit{OnlyRole}.

\paragraph{Results} \autoref{tab:masked_prediction_results} shows the results of our
experiments. We find all of our settings to outperform the
\textit{No-Information} performance-floor (43.5\%) as expected; 
 both \textit{Agreed} (50.3\%) and \textit{EntireSheet}
(47.6\%) outperform the more comparable \textit{Character-Summary}
baseline (44.9\%). This difference is even greater under the
\textit{OnlyRole} setting, where baseline performance decreases but
our \textit{EntireSheet} accuracy jumps to 56.5\%.  This validates our
assertion that \textsc{CHIRON}-based character-representations provide
a tangible benefit on downstream tasks compared to existing
summary-based representations, without any additional training.



We can further improve performance by modifying our \textit{Agreed}
setting to compute agreement exclusively between our Knowledge and
Physical categories. This \textit{Agreed:Knowledge+Physical} setting,
with \textit{OnlyRole}, results in an accuracy of 58.5\%, a full
13.6\% higher than the best performing baseline. The optimal
combination of character-sheet categories would likely change
depending on the downstream task, and we consider it a strength of
\textsc{CHIRON} that it provides this flexibility. 

\section{Story Analysis by Character Density}
\label{sec:story-analys-char}

In addition to being useful for downstream tasks, we wish to show that
\textsc{CHIRON} is useful as a means of automatically analyzing
stories from a variety of sources. We create a metric,
\textit{density}, and show that it correlates with human judgements of
character-centricity before applying it to a varied set of human and
LLM-generated story datasets.


We define \textit{density} as the number of sentences in a finished
character sheet divided by the number of sentences in the provided
story, averaged across all story-character combinations. Let~$S$ denote
the set of story-character combinations, $C$~is the set of categories
(e.g.,~Dialogue, Goals), and $\chi(s, c)$ is the character sheet
category $c$ given a story-character:
\begin{equation*}
\text{density}=\frac{1}{|S|} 
\sum_{s \in S} \sum_{c \in C} \frac{\text{\# sentences in } \chi(s, c)}{\text{\# sentences in story of } s}
\end{equation*}
We interpret this metric as a way of measuring the extent to which
characters are the focus and driving force behind a story, or in other
words a story's \textit{character-centricity}.

We can also compute density for each category in \textsc{CHIRON},
quantifying the extent to which a given aspect of characterization is
present in a story:
\begin{equation*}
\text{density}_c = \frac{1}{|S|} \sum_{s \in S} \frac{\text{\# sentences in } \chi(s, c)}{\text{\# sentences in story of } s}
\end{equation*}


\paragraph{Correlation with Human Judgments}
We evaluate our \textit{density} metric against human judgments on
stories produced by DOC, Re$^3$ and their variants
\citep{yang_doc_2023, yang_re3_2022}. They are state-of-the art
models, exhibiting traditional narrative style. Moreover, their
stories are relatively short (averaging 2k fewer words than our
STORIUM set) and thus amenable to human review.

The Re$^3$ framework introduced a prompting-based method for drafting,
rewriting and editing generated stories \citep{yang_re3_2022}. DOC
built on this framework by adding a more detailed plot outline and a
controller that attempts to match generated stories to the
outline. `No-Control' and `NO-Outline' refer to settings without the
controller and the detailed outline respectively. Both Re$^3$ and DOC
also have an `Interactive' setting, where humans are involved in the
story-planning process \citep{yang_doc_2023}. We collated the output
of their systems, a total of 100 stories across 6 settings, which we
annotated using a similar Prolific setup to our entailment-classifier dataset.

Annotators were asked to rate (from 1--5) a story across five
dimensions meant to encapsulate how well-defined characters are in the
story and how central to the plot: \textbf{Roles} (Does the story have
characters with explicitly defined roles?); \textbf{Traits} (Do the
characters have well-defined traits and relationships?);
\textbf{Actions} (Do their actions play a role in the story?);
\textbf{Physical/Personality} (Do the characters have defined
physical/personality attributes?); and \textbf{Dialogue} (Do the
characters have defined manners of speaking?). We sum up these scores
to give us a singular metric of character importance  which we
correlate with our \textsc{CHIRON}-based \textit{density} metric.

We find that our automated  \emph{density} metric correlates highly with human judgements of character centricity in stories, achieving a   Pearson correlation of~0.753 (\mbox{$p < 0.01$}).  
This allows us to use \textsc{CHIRON} as an
analysis tool on datasets where human annotations would be expensive
or intractable. We can also determine the correlations between
$\text{density}_{\text{dialogue}}$ and
$\text{density}_{\text{physical/personality}}$ and their respective
questions in our annotation task. We find a dialogue correlation of
0.362 ($p < 0.01$) and a physical/personality correlation of 0.440 ($p < 0.01$), further
justifying that the categories within \textsc{CHIRON} are useful for
deeper analysis.

\paragraph{Comparing Stories via Character Density} We  briefly show
an example of how \textsc{CHIRON}'s density metric can be used to
differentiate between different stories and styles. We compute our
density metric across the following datasets, for the most common character in each story:

\textbf{STORIUM} \citep{akoury_storium_2020}: We expect to find a high
density from STORIUM stories due to the extreme character focus
inherent to the way the stories were created (almost akin to
role-playing). We report density based on all-snippets by default as
it is the most comparable between datasets, but we also report density
on the \textit{OnlyRole} setting. 

\textbf{New Yorker TTCW} \citep{chakrabarty_art_2024}: this dataset is
derived from New Yorker stories and LLM-generated stories from a
premise. \citet{chakrabarty_art_2024} showed that the New Yorker
stories were much more creative (as measured by the TTCW\footnote{TTCW is a short for Torrance Test of Creative Writing.} metrics) than
the LLM-generated stories, and were in particular more original in
their Form, Thought, Theme and Content. We expect to find low
density from the New Yorker stories due to their differences from
conventional narratives (a few stories do not even have named main
characters, for example).

\textbf{DOC stories} \citep{yang_doc_2023, yang_re3_2022}: We compare
the automatically generated stories from the following settings:
\textit{DOC}, \textit{DOC-nocontrol}, \textit{DOC-nooutline},
\textit{DOC-interactive}, \textit{Re$^3$},
\textit{Re$^3$-interactive}.

The results in \autoref{tab:density_combined_results} corroborate our
expectations. In particular, the more `creative' narrative structures
of human-written New Yorker stories \citep{chakrabarty_art_2024}
produced a low \textit{density} score, while the character-focused
Storium-OnlyRole setting produced the highest. Results for DOC
\citep{yang_doc_2023} and Re$^3$ \citep{yang_re3_2022} are
harder to interpret, but seem to indicate higher character-centricity
in Re$^3$ stories than DOC ones and higher character-centricity in
stories humans edited (the `Interactive' settings).  Our
\textit{density} metric should not be seen as a score for  story `quality' or
`goodness'. Instead,  it should be used to compare styles of
stories or story generation systems.

\begin{table}[t]
{\small
\centering
\begin{tabular}{llc}
\toprule
\multicolumn{1}{l}{\textbf{Source}} & \multicolumn{1}{l}{\textbf{Setup}} & \multicolumn{1}{c}{\textbf{Density}} \\
\midrule
New Yorker	& New Yorker	& 0.648 \\
DOC	& No-Control	& 0.746 \\
DOC	& DOC	& 0.810 \\
DOC	& Interactive	& 0.840 \\
DOC	& No-Outline	& 0.922 \\
New Yorker	& GPT4	& 1.037 \\
Re$^3$	& Re$^3$	& 1.046 \\
Re$^3$	& Interactive	& 1.058 \\
New Yorker	& Claude	& 1.096 \\
Storium	& Storium	& 1.254 \\
New Yorker	& GPT3.5	& 1.279 \\
Storium	& Storium-OnlyRole	& 1.668 \\
\bottomrule
\end{tabular}
\caption{\label{tab:density_combined_results} \textsc{CHIRON}-based
  density metric computed across different data sources and setups,
  sorted by density. We find human-written New Yorker stories to have
  the lowest density, potentially due to the more creative narrative
  structures referenced in \citet{chakrabarty_art_2024}. In contrast,
  our Storium-OnlyRole setting has the highest density by a wide
  margin, likely explained by the extremely character-driven source
  material \citep{akoury_storium_2020}} }
\end{table}

\section{Conclusion}
\label{sec:conclusion}

In this work we introduce \textsc{CHIRON}, a novel character-sheet
representation for long-form narratives consisting of a zero-shot
Generation Module (in the form of structured question-answering) and a
Validation Module composed of automated reasoning and a trained
entailment-model. Out-of-the-box LLMs frequently ($32.6\%$ of
sentences) produce character descriptions inconsistent with the story,
but our Validation Module can differentiate with $0.930$~precision
between consistent and inconsistent statements.
%
%
%
On the downstream task of masked-character prediction, \textsc{CHIRON}
outperforms (11.6\% gain) a comparable summary-based baseline, which
suggests that the representation would be useful for other
character-centric tasks. Finally, we show that \textit{density}, an
automated metric derived from \textsc{CHIRON}, correlates with human
judgements of character-centricity and can be used to differentiate
story styles.  In future work, we plan to leverage \textsc{CHIRON} to
generate more character-driven stories.

\section*{Limitations}

While the character sheet structure selects for specific types of information, determining what is an `interesting' statement about a character is an open question and one that is hard to answer without knowledge of the entire story. Whether a character holds an object may be only relevant short-term, for example, but could also be crucially important when they pull it out in the final act.

A follow-up study could also better leverage information readers gain across the narrative as context for the current snippet. For example, many snippets in a story will contain story-specific vocabulary or established character relationships, but do not explicitly explain them as they were introduced previously. We only showed annotators one snippet of story at a time, but future work could relax this requirement and ask annotators to read the entire story up to the snippet of interest.

We also focus exclusively on English-language stories in this work;
future research should expand our ontology and annotations to other
languages and explore using other language models pretrained on more
non-English data.

\section*{Ethics Statement}

Any automated system for text generation has the potential to create harmful content. As such, we aimed to avoid potentially inappropriate domains by leveraging the STORIUM dataset's filters. We also took proper precautions to allow crowdsourced workers to withdraw from the task at any point.

It is also possible that abstract character representations (e.g. \textit{`X character is a member of Y group'}) could lead to stereotypical generations when used for story-generation purposes. While we hope that encouraging more complex and rich character representations makes such stereotypes less likely, this is an open question for future research. Such work could also explore the potential for bias in the statement-generation step, where a similar bias may cause models to fill character-sheets with incorrect statements about a character due to stereotypes on mentioned attributes.

\section*{Acknowledgments}

We gratefully acknowledge the support of the UK Engineering and Physical Sciences Research Council (grant EP/L016427/1),
%
\bibliography{anthology,StoryGeneration_Initial_Lit_Review,LitReviewStoryGenerationDatasets,LiteraryWorkForUnderstandingCharacters,ModelsToCite,MiscellaneousBackground,LitReviewCharacterFocusedWork}

\appendix

\section{STORIUM Filtering}
\label{sec:appendix_storiumfiltering}

We filter stories for the following attributes: \textit{play-style} $\in$ [`writing', `standard'], \textit{speed} $!=$ `Hardcore', \textit{suspended} $=$ \textbf{False}, \textit{is\_nsfw} $=$ \textbf{False}, \textit{\# Words} $> 1,000$, \textit{Avg. Words per Entry} $>30$, \textit{Language} $=$ \textbf{English}. We find this filters for English stories in a standard narrative style, and removes potentially offensive content.

For the purposes of our annotations we only select specific entries where \textit{\# Words in Entry} $>50$ to increase the likelihood of character-focused content, and shorter overall works where \textit{\# Words} $< 10,000$ to better allow for future story-long annotations. This last filter removes around 500 stories from consideration, which may be useful for future work. This gives us a dataset of 529 stories which we sample from for entailment dataset. 

Entries are also further split to give us an average word count of 337; we make splits along paragraph lines to maintain comprehensibility as much as possible.

Finally, for the purposes of masked-character-prediction we also filter for \textit{finished} stories (defined as the last scene having \textit{is\_ended}$=$\textbf{True} and \textit{is\_final}$=$\textbf{True}. We also filter for stories that contain a scene-entry with the three most common characters listed by name.

\section{Annotation Collection and Examples}
\label{sec:appendix_examples}

Annotators were collected through Prolific, filtering for: living in the US or UK, and English as a Primary Language. All annotators were paid a rate of at least £9 an hour. 

\autoref{tab:example_annotations} contains a sample annotation for each value in our 1-5 scale, as well as a justification as explanation.

\begin{table*}[h]
\centering
\begin{tabular}{p{0.07\textwidth} p{0.40\textwidth} p{0.15\textwidth} p{0.05\textwidth} p{0.18\textwidth}}
\hline
\textbf{\footnotesize Character} & \textbf{\footnotesize Snippet} & \textbf{\footnotesize Statement} & \textbf{\footnotesize Rating} & \textbf{\footnotesize Justification} \\
\hline
{\footnotesize Holly} & {\footnotesize "Of all the people here likely to have magic, your the least likely. I probably have more magic in my little finger than you on your whole body." Holly said this with confidence in her voice but she was really thinking about if there was someone who could just know someone else had magic. 
Taking a deep breath she sat next to the fire to try and dry off.} & {\footnotesize Based on the given story section, we can infer the following physical descriptions of Holly: \textit{[Llama-2 13b]}} & {\footnotesize 1} & {\footnotesize This statement makes no claims about Holly, and therefore cannot entail from the snippet.} \\
\hline
{\footnotesize Cherish Nicole} & {\footnotesize Cherish was startled by his comment and moved her book to her lap in a rush. 
"Uh, well, no I wasn't taught that. I'm sorry, I didn't know I was being rude.", she said quickly, not meeting Gregory's eyes. She didn't want to give potential friends the wrong impression, darn it. She makes a mental note to leave the textbooks in the room next time. "What are toastmasters?", she asks, completely confused.} & {\footnotesize She uses simple sentences and avoids complex vocabulary when she is unsure of a topic. \textit{[Mistral-7b]}} & {\footnotesize 2} & {\footnotesize Cherish does speak in simple sentences in this snippet while unsure of a topic, but there is no indication that she avoids complex vocabulary and instead seems to ask about terms she does not understand.} \\
\hline
{\footnotesize Dr F. Leiber} & {\footnotesize "I'd be obliged if you'd employ that blade on my restraints." said Dr Lieber as he pulled and strained at the bindings that prevented him from sitting fully upright.
"Only, I beg you, leave my hands unharmed that I may continue my practice."
Turning to look at Professor Doyle, Lieber asks "Are you able to barricade the door with that bed Doyle? Buy us some time."} & {\footnotesize He pleads for help in barricading the door to buy them some time. \textit{[Mistral-7b]}} & {\footnotesize 3} & {\footnotesize While it is true that Leiber asks Doyle to barricade the door to buy some time, it's unclear if Leiber \textit{pleads for help} as they seem to ask calmly.} \\
\hline
{\footnotesize Kaye Madison} & {\footnotesize Kaye shifted her feet nervously. "That's one way of saying it, I suppose. It's not so much funny as it is uncommon, I guess." She readjusted her bag on her shoulder and took a sip of tea. "Still, it's weird to be all friendly and stuff with strangers, right? What if I was some kind of a murderer?"} & {\footnotesize Based on this story section, we can infer that Kaye speaks in a hesitant and self-deprecating manner. \textit{[Llama-2 13b]}} & {\footnotesize 4} & {\footnotesize This statement seems generally true based on the snippet. However, one small complication is if Kaya's comment about potentially being a murderer counts as self-deprecating. It is also unclear if Kaya always speaks in this manner.} \\
\hline
{\footnotesize Anastasia Howe} & {\footnotesize "GAH!" Anastasia screams as she wakes up to a snake curled up. She had about 1 second to take in the new scenery. She then jumped up and away from the snake. She can't fight! Yet again she can't work with animals. She jumped away, uncertain of what the snake might do to her. "Help." She yelped weakly.} & {\footnotesize Anastasia appears to be easily frightened, as evidenced by her scream and immediate jump to escape when she wakes up to a snake. \textit{[GPT-4]}} & {\footnotesize 5} & {\footnotesize The claim and its underlying reasoning all entirely entail from the snippet, and there are no ambiguities.} \\
\hline

\end{tabular}
\caption{\label{tab:example_annotations}
Example annotations and their justifications. While the rest of the table comes directly from our dataset, the justifications are author-provided included for better comprehension.}
\end{table*}

\section{Annotation Interface}
\label{sec:appendix_annotation}

\begin{figure*}[h]
\centering
\includegraphics[width=\textwidth]{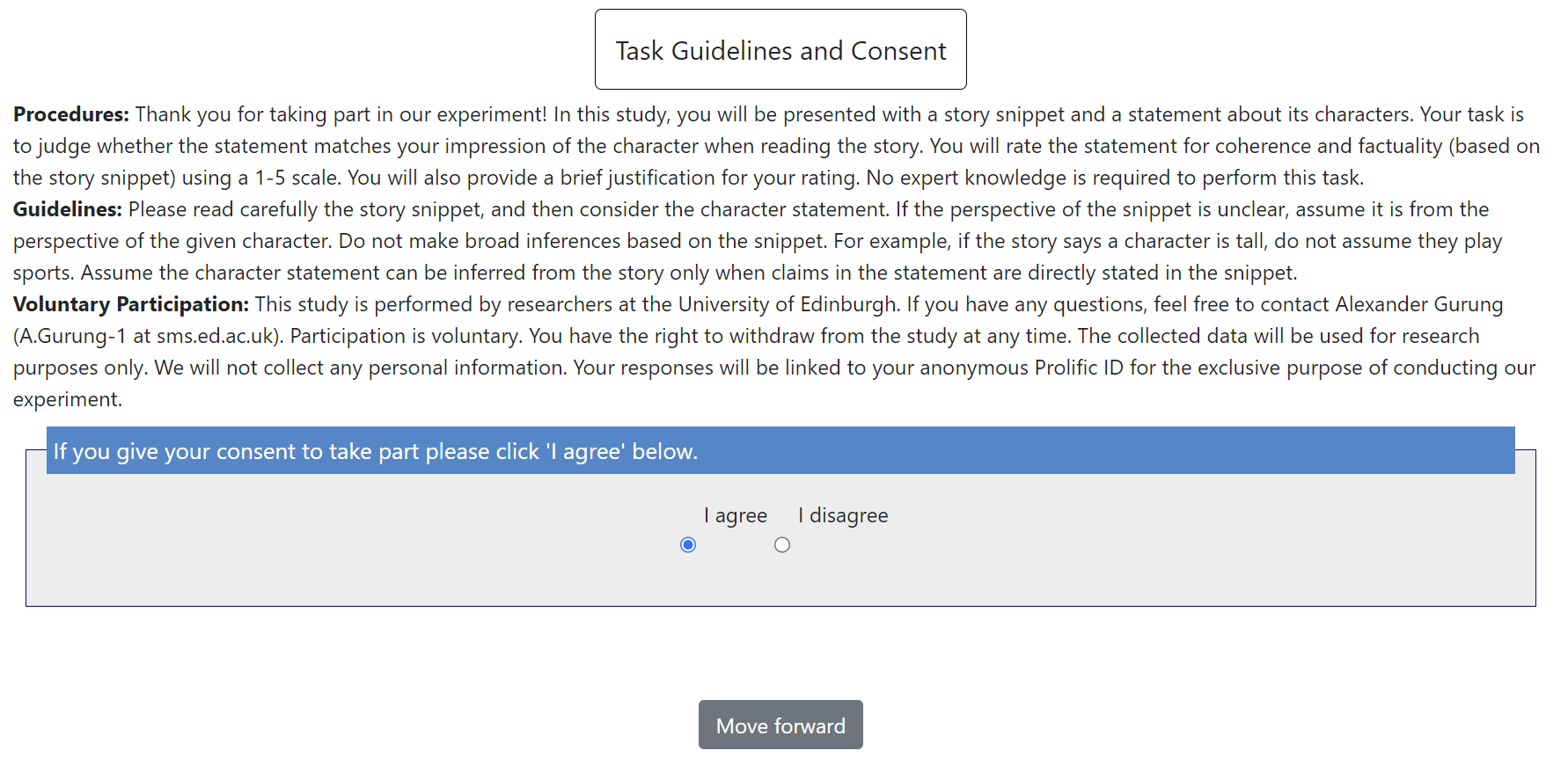}
\caption{\label{fig:potato_guidelines}
Guidelines and consent for task shown to Prolific annotators.
}
\end{figure*}

\begin{figure*}[h]
\centering
\includegraphics[width=\textwidth]{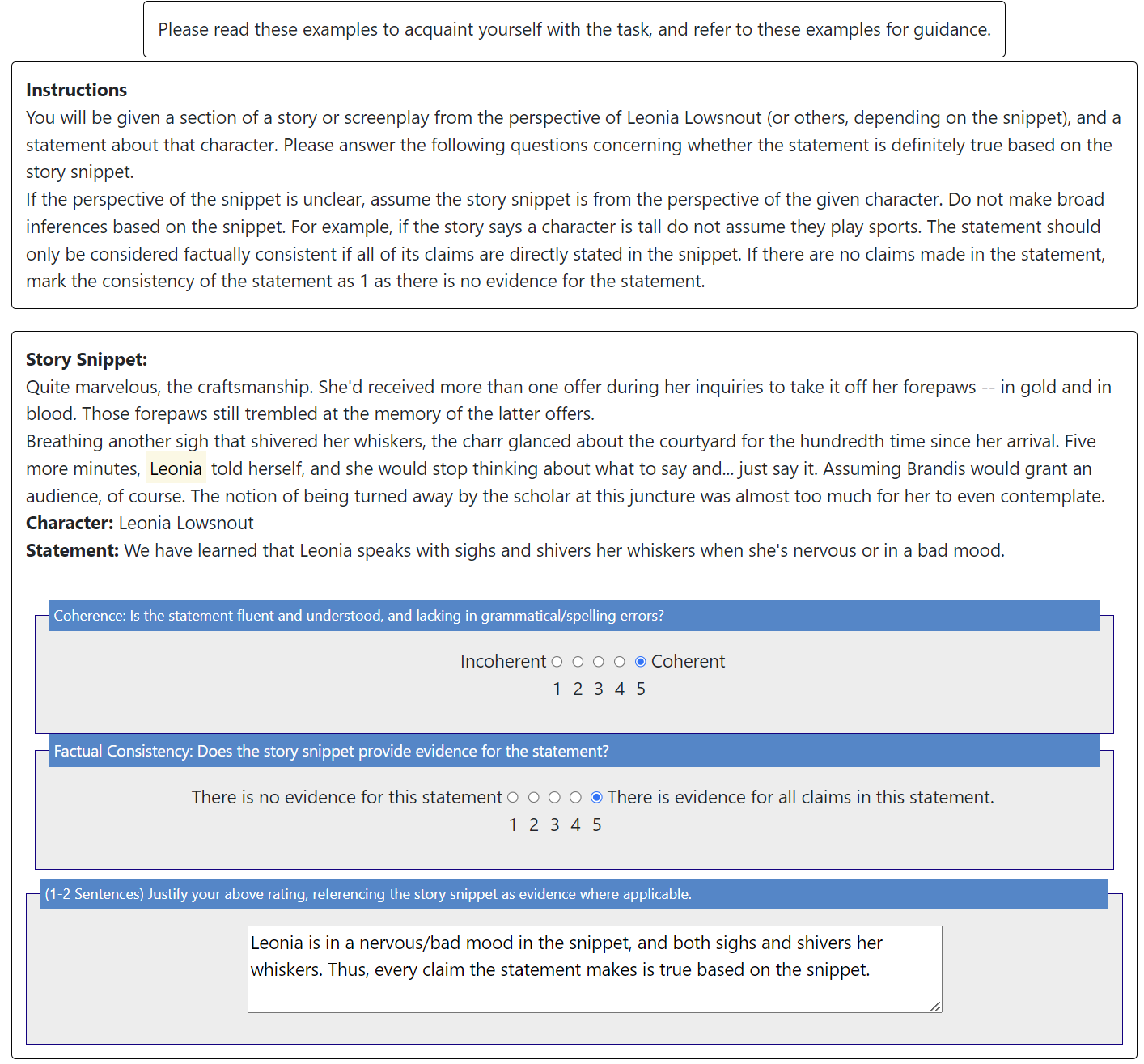}
\caption{\label{fig:potato_instructions}
Instructions for annotation task shown to Prolific annotators, including one of the provided examples.
}
\end{figure*}

\begin{figure*}[h]
\centering
\includegraphics[width=\textwidth]{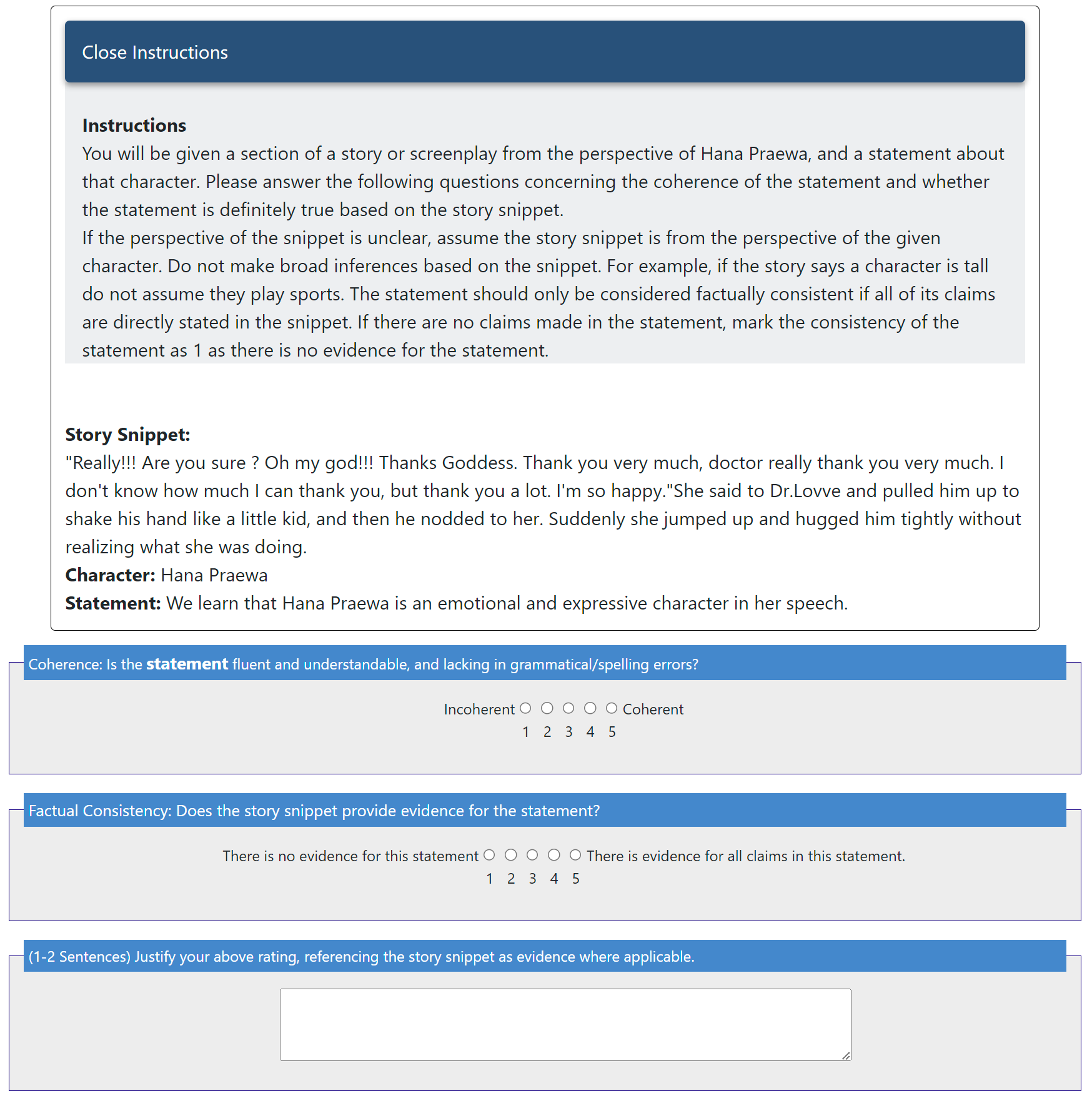}
\caption{\label{fig:potato_example}
Example of annotation task shown to Prolific annotators.
}
\end{figure*}

\autoref{fig:potato_guidelines} (Guidelines \& Consent), \autoref{fig:potato_instructions} (Instructions), and \autoref{fig:potato_example} (Example Annotation) provide screenshots of the annotation task we provided Prolific crowdworkers.

\section{Full Character Sheet Examples}
\label{sec:appendix_full_character_sheet_examples}

\autoref{tab:full_character_sheet_p1} and \autoref{tab:full_character_sheet_p2} contain a full example character-sheet (over the course of an entire story) before and after the Validation Module.

\begin{table*}[h]
\centering
\begin{tabular}{p{0.95\linewidth}}
\hline
{\large \textbf{Physical/Personality} }\\
\textbf{What, if any, physical descriptions of this character are in this snippet?} \\

{\footnotesize \colorbox{yellowgreen}{Nadia is described as watching both men silently. She pulls her tattered, too-big sweatshirt tighter around herself. }} \\
{\footnotesize \colorbox{yellowgreen}{Nadia is a woman with the ability to glance over her shoulder. She deflates slightly. She sees that no men have followed}} \\
{\footnotesize \colorbox{yellowgreen}{her.} No other physical descriptions are provided in this snippet.} \\
{\footnotesize \colorbox{yellowgreen}{Nadia is described as pushing her way forward. She has companions who scoff. } There is no physical description of her appearance given in this snippet.} \\
{\footnotesize Nadia is described as staring in horror. \colorbox{yellowgreen}{She bites her lip. She glances towards Ian. } She shouts to the natives. \colorbox{yellowgreen}{She pulls out a digital camera. She is shocked when the natives react positively to the camera. } She is dragged down from the rooftop by a woman. There is no detailed physical description of Nadia in this snippet.} \\
{\footnotesize \colorbox{yellowgreen}{Nadia bit her lip. } She glanced up at the men. She gestured in the direction of the cell tower.} \\
\hline
\textbf{What, if any, descriptions of this character's personality are in this snippet?} \\

{\footnotesize \colorbox{yellowgreen}{Nadia is quiet. She has not spoken since the plane crash. She is observant. } She notices the approaching argument between Jacob and Jayson. \colorbox{yellowgreen}{She is cautious and concerned about their safety. She suggests they seek shelter in the cave} \colorbox{yellowgreen}{during an impending storm. She is emotionally drained. }} \\
{\footnotesize \colorbox{yellowgreen}{Nadia is angry.} She expects men to follow her willingly. She is facing reality. \colorbox{yellowgreen}{She is introspective. }} \\
{\footnotesize \colorbox{yellowgreen}{Nadia is assertive. Nadia is apologetic for her friends' behavior. Nadia is determined to obtain information. She offers him}  \colorbox{yellowgreen}{the opportunity to accompany them off the island. However, she is firm in her demand for full disclosure from him. }} \\
{\footnotesize \colorbox{yellowgreen}{Nadia is resourceful. She is brave. } She is willing to take risks, as shown when she calls out to the natives from the rooftop. \colorbox{yellowgreen}{She is also compassionate and caring. } She is culturally curious and open-minded.} \\
{\footnotesize \colorbox{yellowgreen}{Nadia is determined and brave. She knows about the cell tower and offers this information to the men. Nadia is }} \\
{\footnotesize \colorbox{yellowgreen}{independent and strong-willed. She chooses to stay on the island instead of leaving with the men. } She is aware of the potential danger and the consequences of Ian and his wife's actions.} \\

\hline
{\large \textbf{Knowledge} }\\
\textbf{What, if any, factual information is given about this character in this snippet?} \\
{\footnotesize \colorbox{yellowgreen}{Nadia is a survivor of a plane crash. She has been silent since the accident. She wears a tattered sweatshirt. She is } \colorbox{yellowgreen}{emotionally drained. She suggests seeking shelter in a cave. }} \\
{\footnotesize Nadia is the name of the character. She is the one who glances over her shoulder in fear. \colorbox{yellowgreen}{She is an author. }} \\
{\footnotesize \colorbox{yellowgreen}{Nadia is a member of a group. She apologizes for her friends' behavior towards Ian. } She offers him the opportunity to accompany them off the island in exchange for information. \colorbox{yellowgreen}{She is assertive. } She takes charge of the situation. \colorbox{yellowgreen}{She has a backpack with her. }} \\
{\footnotesize \colorbox{yellowgreen}{Nadia is a character who witnesses Ian falling off a rooftop. Nadia is confronted by hostile natives. She is resourceful and} \colorbox{yellowgreen}{quick-thinking. } She uses a digital camera to diffuse the situation. \colorbox{yellowgreen}{The natives react positively to the camera. } They allow Nadia to be helped down from the rooftop. \colorbox{yellowgreen}{Nadia shows concern for Ian. } Nadia apologizes when he is injured.} \\
{\footnotesize \colorbox{yellowgreen}{Nadia is stranded on an island. She hates two men, Jayson and Jacob. She knows of a cell tower nearby. } She suggests they use it to contact someone for help. \colorbox{yellowgreen}{She decides to stay on the island. } She is certain she is making the right choice.} \\

\hline
\textbf{What, if any, information has this character learned in this snippet?} \\
{\footnotesize Nadia has learned that she has valuable input in decision-making situations. She has observed the potential for conflict between Jacob and Jayson. She has suggested a course of action to avoid it. \colorbox{yellowgreen}{She is aware of the approaching storm. }} \\
{\footnotesize \colorbox{yellowgreen}{Nadia has realized that the men she is with are not willing to follow her like the characters in her novels. She feels}} \\
{\footnotesize \colorbox{yellowgreen}{disappointed.}}
{\footnotesize \colorbox{yellowgreen}{Nadia has learned that Ian is in possession of valuable supplies. She learns that her friends are treating Ian }} \\
{\footnotesize \colorbox{yellowgreen}{unfairly.}}  \\
{\footnotesize \colorbox{yellowgreen}{Nadia learns that the natives of the island are fascinated by the camera. } Using it as a tool, she gains their trust. She is able to safely descend from the rooftop.} \\
{\footnotesize \colorbox{yellowgreen}{Nadia learns that there is a cell tower nearby. } Nadia realizes that the men, Jayson and Jacob, could use it to contact someone for help. \colorbox{yellowgreen}{She realizes that she is not leaving the island. } She will stay there to live in the world of her books.} \\
\hline

\end{tabular}
\caption{\label{tab:full_character_sheet_p1}
Sample part (Physical/Personality and Knowledge) of character sheet concatenated across an entire story, all text is from the Generation Module and highlighted text is that which passed the Validation Module.}
\end{table*}

\begin{table*}[h]
\centering
\begin{tabular}{p{0.95\linewidth}}
\hline
{\large \textbf{Dialogue} }\\
\textbf{What, if anything, have we learned about how this character speaks from this snippet?} \\

{\footnotesize \colorbox{yellowgreen}{Nadia has a quiet and hesitant speaking voice. She clears her throat before speaking. She has gone without using her voice} \colorbox{yellowgreen}{for an extended period. }} \\
{\footnotesize \colorbox{yellowgreen}{Nadia thinks aloud. }} \\
{\footnotesize \colorbox{yellowgreen}{Nadia speaks assertively. Nadia interrupts conversations when necessary. She apologizes for her friends' behavior. She} \colorbox{yellowgreen}{stands firm on the need for information. She uses direct language. } She makes clear the conditions for allowing Ian to accompany them.} \\
{\footnotesize Nadia speaks with urgency when demanding answers. \colorbox{yellowgreen}{She uses exclamations such as "Screw it." } She communicates effectively with the natives.} \\
{\footnotesize \colorbox{yellowgreen}{Nadia speaks with determination. She shouts to get the men's attention. } She communicates effectively about the cell tower's location. \colorbox{yellowgreen}{She is assertive in expressing her decision to stay on the island. }} \\
\hline

{\large \textbf{Plot} }\\
\textbf{What, if any, goals does this character gain in this snippet that they wish to accomplish in the future?} \\

{\footnotesize \colorbox{yellowgreen}{Nadia's goal in this snippet is to seek shelter from the approaching storm. }} \\
{\footnotesize Nadia desires to have people willingly follow her in reality. Just as they do in her novels.} \\
{\footnotesize \colorbox{yellowgreen}{Nadia's goal in this snippet is to obtain necessary information from Ian. She apologizes for her friends' behavior. } She offers him the opportunity to accompany them. To achieve this, she convinces Ian to share his supplies.} \\
{\footnotesize \colorbox{yellowgreen}{Nadia's initial goal is to protect Ian. } Nadia demands answers from the men who pushed him off the rooftop. However, when she encounters the natives, her goal shifts to finding a way to communicate and connect with them. \colorbox{yellowgreen}{She discovers that the natives are fascinated by the camera. } Her new goal is to use the camera to build a relationship with the natives.} \\
{\footnotesize Nadia's goal in this snippet is to help the men contact someone off the island. \colorbox{yellowgreen}{She stays behind. }} \\

\hline
\textbf{What, if any, goals does this character complete in this snippet?} \\

{\footnotesize \colorbox{yellowgreen}{Nadia expresses her opinion to Jacob and Jayson. Nadia suggests seeking shelter in the cave during an approaching storm.} \colorbox{yellowgreen}{She initiates communication after a long period of silence following the plane crash. }} \\
{\footnotesize \colorbox{yellowgreen}{Nadia glances over her shoulder. She is disappointed that neither man follows her. She thinks angrily about how things} \colorbox{yellowgreen}{would be different in her novels. } No goals are completed in this snippet for Nadia.} \\
{\footnotesize \colorbox{yellowgreen}{Nadia convinces Ian to share his supplies with her group. }} \\
{\footnotesize Nadia manages to save herself from the natives. \colorbox{yellowgreen}{She uses a digital camera to distract them. She discovers that the natives are} \colorbox{yellowgreen}{fascinated by the camera. }} \\
{\footnotesize \colorbox{yellowgreen}{Nadia reveals the existence of a cell tower to Jayson and Jacob. Despite the danger, she decides to stay on the island. }} \\

\hline
\textbf{How, if at all, does this character's internal motivations change in this snippet?} \\

{\footnotesize \colorbox{yellowgreen}{Nadia, who had been silent since the plane crash, feels the need to speak up. } She speaks up when she sees the men arguing about entering the cave. \colorbox{yellowgreen}{She expresses her concern about the approaching storm. She suggests they} \colorbox{yellowgreen}{seek shelter in the cave. } Despite her emotional exhaustion, she takes the initiative.} \\
{\footnotesize Nadia's internal motivation shifts from expecting the men to follow her. \colorbox{yellowgreen}{She compares the situation to her novels. } She implies a desire for excitement and adventure in her real life.} \\
{\footnotesize \colorbox{yellowgreen}{Nadia feels apologetic for her friends' behavior towards Ian. Nadia remains determined to obtain necessary information. } She offers him the opportunity to join them until they leave the island. She insists on receiving all the information he possesses in return.} \\
{\footnotesize Nadia's initial motivation is to protect Ian. Nadia demands answers from the men who pushed him off the rooftop. However, she realizes that the natives are not hostile towards her but rather fascinated by the camera. Her motivation shifts to finding a way to communicate and connect with them. \colorbox{yellowgreen}{She uses the camera as a tool. }} \\
{\footnotesize Nadia's internal motivation shifts from hatred and a desire to be rid of the men. \colorbox{yellowgreen}{She develops a desire to stay on} \colorbox{yellowgreen}{the island and live in the world of her books. She makes this decision despite the danger and potential consequences. }} \\

\hline
\end{tabular}
\caption{\label{tab:full_character_sheet_p2}
Sample part cont. (Dialogue and Plot) of character sheet concatenated across an entire story, all text is from the Generation Module and highlighted text is that which passed the Validation Module.}
\end{table*}

\section{Automated Reasoning Steps}
\label{sec:appendix_automatedreasoning}

\subsubsection{In-Context-Learning}

Due to the relatively long nature of our snippets, ICL is not naturally applicable. However, we can extract from our overarching annotation task two sub-tasks that do not require the snippet: 

\textbf{Ambiguity:} does this statement contain references to unspecified objects? Example: \textit{He ran away from them.} Too much ambiguity may make a statement impossible to entail from the snippet. 

\textbf{Informativeness:} does this statement provide any information about the character? Example: \textit{There were no descriptions of Kelly.} Statements that do not make any claims do not entail from the snippet.

We generate responses starting with \textit{Yes} or \textit{No} followed by a short sentence justifying the answer. We find that naively using the boolean values as a filtering step for the classifier is too aggressive and has a low recall, largely due to many statements being slightly ambiguous. The justification is thus useful to further explain the rating so the downstream classifier can choose how heavily to consider it. Examples are given in \autoref{tab:example_icl_statements}.

\begin{table*}[t]
\centering
\begin{tabular}{l p{0.42\textwidth} p{0.42\textwidth}}
\hline
\textbf{\footnotesize Type} & \textbf{\footnotesize Statement} & \textbf{\footnotesize Reasoning} \\
\hline
{\footnotesize Ambiguity} & {\footnotesize However, before he can carry out his plan, she is shot from behind and collapses into his arms.} & {\footnotesize Yes, the statement is ambiguous as it does not specify who "she" is.} \\
{\footnotesize Ambiguity} & {\footnotesize Skótos, also known as Scott Mavrogiannis, is a demi-god.} & {\footnotesize No, the statement is unambiguous in stating that Skótos is a demi-god.} \\
{\footnotesize Informative} & {\footnotesize She is also adaptable, following William into the dark room without complaint.} & {\footnotesize Yes, the statement describes Sharon as adaptable} \\
{\footnotesize Informative} & {\footnotesize No other physical descriptions are provided in this snippet.} & {\footnotesize No, the statement does not provide any new physical descriptions of Jameson Jackson.} \\
\hline
\end{tabular}
\caption{\label{tab:example_icl_statements}
Example ICL reasoning statements and the corresponding automatically generated responses. A response of `Yes' for Ambiguity questions implies the statement is too ambiguous to be understood, and likely should not be included in a character sheet. A response of `No' for Informative questions implies the statement does not provide any new information, and likely should not be included in a character sheet.}
\end{table*}

\subsubsection{Chain of Thought}

We mimic chain of thought prompting by asking the model two short-answer questions that annotators implicitly answered while labeling. First, we ask for the section of the snippet most relevant to the statement. Second, we ask the model to compare and contrast the section of the snippet to the statement. We hypothesize that implicitly asking the model to both identify key phrases and compare them to the statement at the same time is too difficult, and find that guiding the model's reasoning in this way improves performance. We provide examples in \autoref{sec:appendix_cot} \autoref{tab:example_chain_of_thought}.

\section{Chain of Thought Example}
\label{sec:appendix_cot}

\autoref{tab:example_chain_of_thought} contains two examples of our chain-of-thought reasoning steps, to illustrate how they might help determine the validity of a given statement.

\begin{table*}[t]
\centering
\begin{tabular}{p{0.13\textwidth} p{0.38\textwidth} p{0.39\textwidth}}
\hline
\textbf{\footnotesize Statement} & \textbf{\footnotesize Question} & \textbf{\footnotesize Answer} \\
\hline
{\footnotesize She intends to distract any potential threats and lead Ymros to safety through the fire door.} & {\footnotesize What, if any, section of the story snippet is most relevant to the given statement? Provide a brief 1-2 sentence description of this section or "N/A" if there is no relevant section.} & {\footnotesize The section of the story snippet that is most relevant to the given statement is when Rachel suggests going outside through the fire door to distract potential threats and lead Ymros to safety.} \\
 & {\footnotesize In 1-2 sentences, compare the claim the statement makes and the section of story you highlighted in your previous answer. Are there any notable differences? Are all claims made by the statement explicitly supported? If there are no claims, write "N/A".} & {\footnotesize The statement aligns with the section of the story where Rachel plans to use the fire door as a means to distract threats and lead Ymros to safety. Both the statement and the story suggest that Rachel intends to use this strategy to ensure Ymros' safety. No notable differences exist between the two. All claims made by the statement are explicitly supported by the story.} \\

\hline
 
 {\footnotesize Bryn's speech is straightforward and direct.} & {\footnotesize What, if any, section of the story snippet is most relevant to the given statement? Provide a brief 1-2 sentence description of this section or "N/A" if there is no relevant section.} & {\footnotesize Bryn's reaction to Sam's interaction with Mona, specifically when she punches Sam and tears up the card, is the most relevant section to the given statement. In this moment, Bryn's words and actions are straightforward and direct as she expresses her jealousy and frustration towards Mona and Sam.} \\
 & {\footnotesize In 1-2 sentences, compare the claim the statement makes and the section of story you highlighted in your previous answer. Are there any notable differences? Are all claims made by the statement explicitly supported? If there are no claims, write "N/A".} & {\footnotesize The statement that "Bryn's speech is straightforward and direct" is supported by the section of the story where Bryn punches Sam and tears up the card in response to Sam's interaction with Mona. This action demonstrates Bryn's straightforward and direct emotional response to the situation.}  \\
\hline

\end{tabular}
\caption{\label{tab:example_chain_of_thought}
Example Chain of Thought Reasoning}
\end{table*}

\section{Classifier Training}
\label{sec:appendix_classifier_details}

We used Mistral 7B Instruct v0.2 \citep{jiang_mistral_2023} as our base, and used Huggingface's TRL library for all training. LoRA hyper-parameters: $r=16, \alpha=32, \text{dropout}=0.1$, target-modules=["gate\_proj", "down\_proj", "up\_proj", "q\_proj", "v\_proj", "k\_proj", "o\_proj"]. Other hyper-parameters: learning\_rate$=2e-5$, batch-size=$1$ (with gradient-accumulation-steps=$4$), epochs=$4$ (and best model selected by validation precision). Further details will be shared with the code.

All experiments and training was performed on two NVIDIA A100 GPUs.

\begin{table*}[t]
\centering
\begin{tabular}{p{0.17\textwidth} p{0.05\textwidth} p{0.05\textwidth} p{0.12\textwidth} p{0.1\textwidth} p{0.1\textwidth} p{0.1\textwidth}}
\hline
\textbf{Name} & \textbf{ICL} & \textbf{CoT} & \textbf{Eval-Setup} & \textbf{Precision} & \textbf{Recall} & \textbf{Accuracy} \\
\hline
Zero-Shot All & \cmark & \cmark & $\geq 4$ & 0.747 & \textbf{1.0} & 76\% \\
Zero-Shot ICL & \cmark & \xmark & $\geq 4$ & 0.759 & 0.887 & 72\% \\
Zero-Shot CoT & \xmark & \cmark & $\geq 4$ & 0.747 & \textbf{1.0} & 76\% \\
Zero-Shot None & \xmark & \xmark & $\geq 4$ & 0.793 & 0.972 & 80\% \\
Trained All & \cmark & \cmark & $\geq 4$ & 0.868 & 0.93 & 85\% \\
Trained ICL & \cmark & \xmark & $\geq 4$ & 0.868 & 0.831 & 79\% \\
Trained CoT & \xmark & \cmark & $\geq 4$ & 0.843 & 0.986 & 86\% \\
Trained None & \xmark & \xmark & $\geq 4$ & 0.873 & 0.972 & \textbf{88\%} \\
\hline
Zero-Shot All & \cmark & \cmark & $=5$ & 0.753 & 0.986 & 76\% \\
Zero-Shot ICL & \cmark & \xmark & $=5$ & 0.759 & 0.887 & 72\% \\
Zero-Shot CoT & \xmark & \cmark & $=5$ & 0.753 & 0.986 & 76\% \\
Zero-Shot None & \xmark & \xmark & $=5$ & 0.835 & 0.930 & 82\% \\
Trained All & \cmark & \cmark & $=5$ & 0.930 & 0.746 & 78\% \\
Trained ICL & \cmark & \xmark & $=5$ & \textbf{0.952} & 0.563 & 67\% \\
Trained CoT & \xmark & \cmark & $=5$ & 0.864 & 0.803 & 77\% \\
Trained None & \xmark & \xmark & $=5$ & 0.895 & 0.718 & 74\% \\
\hline
\end{tabular}
\caption{\label{tab:all_classifier_performance}
Classifier performance by training data and evaluation type. Zero-Shot refers to untrained Mistral 7B Instruct v0.2. ($\geq 4$ v. $\geq 4$) refers to converting \textbf{both} model predictions and dataset labels from 1-5 to booleans via $\geq 4$. In contrast, ($= 5$ v. $\geq 4$) refers to converting dataset labels the same way but only taking model predictions as \textit{True} if the prediction $=5$. We generally value higher precision, but the best performing precision model (Trained ICL) has much worse accuracy and recall. We take the reasonable performance of Zero-Shot models to indicate the quality of our prompts, but note from the high recall+low precision that this setup over-predicts positive labels. All reported numbers are from our test set (out of an 80-10-10 split).}
\end{table*}

\section{Prompt Templates}
\label{sec:appendix_prompt}

\autoref{tab:initial_generating_prompt_template} shows the prompt template for our Generation Module, and \autoref{tab:simplifying_generating_prompt_template} shows the template for the simplification step that we pass the output from the initial generation through. We added this simplification step after collecting our annotated dataset, and found that it produces both easier to annotate statements and more accurate character sheets. 
Furthermore, by making each claim atomic we ensure our `density' metric uses accurate estimations for the amount of character information.

\autoref{tab:icl_ambiguity_generating_prompt_template} and \autoref{tab:icl_ambiguity_cont_generating_prompt_template} show the template for our `ambiguity' reasoning in our Validation Module. \autoref{tab:icl_informative_generating_prompt_template} and \autoref{tab:icl_informative_cont_generating_prompt_template} show the template for our `informative' reasoning in our Validation Module.

\autoref{tab:cot_generating_prompt_template} shows the template for our Chain-Of-Thought reasoning step in our Validation Module.

\begin{table*}[t]
\centering
\begin{tabular}{p{0.95\textwidth}}

You are a helpful and expert writing assistant. You will be given a section of a story or screenplay. Please answer the following questions about the character learned in this story section, and respond in short paragraph form.

Story Section:

\colorbox{apricot}{[story\_section]}

Please answer the following questions about \colorbox{apricot}{[character]} with short, succinct sentences based on the given story section.

Question: \colorbox{apricot}{[question]} Respond in paragraph form with short, simple sentences with no dependent clauses or transition words.

Answer: \\

\hline
\end{tabular}
\caption{\label{tab:initial_generating_prompt_template}
Prompt for generating initial responses from out-of-the-box LLMs. Highlighted \colorbox{apricot}{[text]} is filled in with the appropriate information prior to generation. Prompts were adjusted to use instruction tokens via their respective model's chat-template.}
\end{table*}

\begin{table*}[h]
\centering
\begin{tabular}{p{0.95\textwidth}}

You are an expert writing assistant helping an author split compound sentences. Please answer the following questions to the best of your ability.

Given the provided sentence, please split all independent clauses into independent sentences and resolve any issues with unclear pronouns or references. Only do this for compound sentences. Every new sentence should make sense on its own. Write them out in paragraph form, one sentence after another.

Sentence: She's curious about a closed door in Maxim's apartment and feels a strong urge to discover what's behind it. 

Split Sentences: She's curious about a closed door in Maxim's apartment. She feels a strong urge to discover what's behind the closed door in Maxim's apartment.

Sentence: Kaluros is determined and focused during battles, using his magic and weapons effectively to defeat his enemies. 

Split Sentences: Kaluros is determined and focused during battles, using his magic and weapons effectively to defeat his enemies.

Sentence: Hassan encountered a crab monster and engaged in a card battle to defeat it. 

Split Sentences: Hassan encountered a crab monster. Hassan engaged in a card battle to defeat the crab monster.

Sentence: She uses imperatives to give orders and asks direct questions to gather information. 

Split Sentences: She uses imperatives to give orders. She asks direct questions to gather information.

Sentence: Bob is easily distracted and forgets about the chase when he notices something outside. 

Split Sentences: Bob is easily distracted. Bob forgets about the chase when he notices something outside.

Sentence: Rachel enters the warehouse to join the baby dragon, defying her initial skepticism. 

Split Sentences: Rachel enters the warehouse to join the baby dragon, defying her initial skepticism.

Sentence: He gives commands to his companions and asks for their assistance. 

Split Sentences: He gives commands to his companions. He asks for his companions' assistance.

Sentence: She explores the Zombear's massive body and climbs on it. 

Split Sentences: She explores the Zombear's massive body. She climbs on the Zombear.

Sentence: Jordan opens the locker to find a locket, a newspaper, and a mysterious photograph. 

Split Sentences: Jordan opens the locker to find a locket, a newspaper, and a mysterious photograph.

Sentence: He is quiet and tosses a gold idol between his hands while they wait for rescue. 

Split Sentences: He is quiet. He tosses a gold idol between his hands while they wait for rescue.

Sentence: \colorbox{apricot}{[sentence]}

Split Sentences: \\

\hline
\end{tabular}
\caption{\label{tab:simplifying_generating_prompt_template}
Prompt (ICL-style) for simplifying responses from out-of-the-box LLMs. We use this to break apart complex claims and make the resulting statements easier to verify in our Verification Module. Highlighted \colorbox{apricot}{[text]} is filled in with the appropriate information prior to generation. Prompts were adjusted to use instruction tokens via their respective model's chat-template.}
\end{table*}

\begin{table*}[t]
\footnotesize
\centering
\begin{tabular}{p{0.98\textwidth}}

You are a helpful and expert writing assistant. Please answer the following questions to the best of your ability.

Question: \colorbox{apricot}{[question\_text]}

Statement: No other physical descriptions of Collins are provided in this snippet. 

Answer: No, the statement is unambiguous in its meaning as there are no claims made.

Question: \colorbox{apricot}{[question\_text]}

Statement: He has a strange glowing key that he uses to open the door to his home. 

Answer: No, the statement is unambiguous in its description of Mustafa's ownership of key and his opening of a door. The pronouns 'he' and 'his' unambiguously refer to Mustafa.

Question: \colorbox{apricot}{[question\_text]}

Statement: These men run away at first sight. 

Answer: Yes, the statement is ambiguous because it doesn't specify who "These men" are, and it is also unclear what "first sight" refers to.

Question: \colorbox{apricot}{[question\_text]}

Statement: They are skittish and afraid of the darkness around their camp. 

Answer: No, the statement is unambiguous about their (Alex's) fear of the darkness.

Question: \colorbox{apricot}{[question\_text]}

Statement: They struggle to stand up. 

Answer: No, the statement is unambiguous about Luis's difficulty standing up.

Question: \colorbox{apricot}{[question\_text]}

Statement: Arjun's primary goal is to resolve this situation. 

Answer: Yes, the statement is ambiguous because we cannot understand Arjun's goal without knowing "this situation".

Question: \colorbox{apricot}{[question\_text]}

Statement: She speaks softly to them, hoping to calm them down. 

Answer: Yes, the statement is ambiguous because it is unclear who "them" refers to, which is necessary to understand the statement.

Question: \colorbox{apricot}{[question\_text]}

Statement: She speaks using colloquial expressions like "C'est la vie" and "TGIF, am I right?". 

Answer: No, the statement is unambiguous in its description of Ping's speaking habits.

Question: \colorbox{apricot}{[question\_text]}

Statement: He is skinny with long legs. 

Answer: No, the statement is unambiguous in its description of Santiago.

Question: \colorbox{apricot}{[question\_text]}

Statement: He opens a brown bottle and drinks the murky liquid inside. 

Answer: No the statement is unambiguous in its description of Jarvis drinking from the brown bottle.

Question: \colorbox{apricot}{[question\_text]}

Statement: Her keen eyes spot Arthur sneaking the coin out from underneath the cup, showing her skills in observation. 

Answer: No the statement is unambiguous in its description of the Bartender's observational skills.

Question: \colorbox{apricot}{[question\_text]}

Statement: He aims to accomplish this by using fire as a means to drive it away, based on his knowledge of myths and lore from various cultures. 

Answer: Yes, the statement is ambiguous as it is unclear what he is trying to accomplish or who/what he is driving aways.

Question: \colorbox{apricot}{[question\_text]}

Statement: Overall, Jake seems to be focused on his social life and maintaining his reputation. 

Answer: No, the statement is unambiguous in its description of Jake's focus.

Question: \colorbox{apricot}{[question\_text]}

Statement: His accomplishes this goal quickly. 

Answer: Yes, the statement is ambiguous as it is unclear what "this goal" refers to. 

[...]
\\

\hline
\end{tabular}
\caption{\label{tab:icl_ambiguity_generating_prompt_template} ICL-prompt: ambiguity. We use this in our Verification Module to determine if a given statement is too ambiguous to verify or do entailment with. Highlighted \colorbox{apricot}{[text]} is filled in with the appropriate information prior to generation. The \colorbox{apricot}{[question\_text]} in this case refers to: \textit{``Is the given statement about \colorbox{apricot}{[character]} ambiguous in a way that makes the meaning unclear? Ambiguities may include, but are not limited to, references to unspecified characters, objects, and actions. If the statement begins with a personal pronoun (e.g. "He" or "She"), assume it refers to \colorbox{apricot}{[character]} and don't count the pronoun towards the ambiguity. Begin your 1-2 sentence response with "Yes" if the statement is too ambiguous to be understood on its own and "No" if the statement makes is unambiguous in its meaning.''}. Prompts were adjusted to use instruction tokens via their respective model's chat-template.. See \autoref{tab:icl_ambiguity_cont_generating_prompt_template} for the rest of the prompt.}
\end{table*}

\begin{table*}[t]
\footnotesize
\centering
\begin{tabular}{p{0.98\textwidth}}

You are a helpful and expert writing assistant. Please answer the following questions to the best of your ability.

Question: \colorbox{apricot}{[question\_text]}

Statement: Collins learned that his father died after the factory accident in 1973. 

Answer: Yes, the statement provides information about what Collins has learned about his father's death.

Question: \colorbox{apricot}{[question\_text]}

Statement: There were no descriptions of Kelly.

Answer: No, the statement does not give you any information about Kelly.

Question: \colorbox{apricot}{[question\_text]}

Statement: She gestures wildly, and speaks with flowerly descriptions.

Answer: Yes, the statement provides information on how Ani speaks.

Question: \colorbox{apricot}{[question\_text]}

Statement: They have a new goal of finding a place to eat.

Answer: Yes, the statement is gives us information about Dr. Alex's new goal.

Question: \colorbox{apricot}{[question\_text]}

Statement: This snippet provides many physical descriptions of Rockefeller.

Answer: No, the statement does not provide any information about Rockefeller.

Question: \colorbox{apricot}{[question\_text]}

Statement: She is an extremely focused and driven individual.

Answer: Yes, the statement is describes the Bartender's personality.

Question: \colorbox{apricot}{[question\_text]}

Statement: Jennifer aims to accomplish this by using fire as a means to drive it away, based on her knowledge of myths and lore from various cultures.

Answer: No, the statement does not make any claims about Arjun as it only makes claims about Jennifer.

Question: \colorbox{apricot}{[question\_text]}

Statement: She expresses a desire to look into the matter further and gather more information, indicating that she is motivated to uncover the truth about the space/time anomaly and its potential impact on the city's population.

Answer: Yes, the statement describes her desire to investigate.

Question: \colorbox{apricot}{[question\_text]}

Statement: He is skinny with long legs.

Answer: Yes, the statement describes his physical appearance.

Question: \colorbox{apricot}{[question\_text]}

Statement: According to this section of the story, Merlin's primary goal is to live to see the night.

Answer: Yes, the statement describes describes Merlin's goal to survive.

Question: \colorbox{apricot}{[question\_text]}

Statement: His internal motivations do not significantly change in this snippet.

Answer: No, the statement does not provide any new information about Mohammed.

Question: \colorbox{apricot}{[question\_text]}

Statement: She recently acquired a gun while searching the trunks abandoned in the cave.

Answer: Yes, the statement tells the reader a fact about her recent activity and a new item she possesses.

Question: \colorbox{apricot}{[question\_text]}

Statement: Here are some physical descriptions of Pedro based on the given story section:

Answer: No, the statement does not provide any information about Pedro.

Question: \colorbox{apricot}{[question\_text]}

Statement: They are a tall warrior.

Answer: Yes, the statement describes their height and profession.

[...]
\\

\hline
\end{tabular}
\caption{\label{tab:icl_informative_generating_prompt_template} ICL-prompt: informative. We use this in our Verification Module to determine if a given statement makes any claims or tells us any information about a character. Highlighted \colorbox{apricot}{[text]} is filled in with the appropriate information prior to generation. The \colorbox{apricot}{[question\_text]} in this case refers to: \textit{``Does this statement give you any novel information concerning \colorbox{apricot}{[character]} or what \colorbox{apricot}{[character]} knows? Novel information may include, but is not limited to, physical descriptions, new information they may have learned, goals they have or actions they have just completed, and descriptions of their speech. Begin your response with "Yes" if the statement gives us any new information and "No" if the statement doesn\'t add to our knowledge/understanding of the character in any way.''}. Prompts were adjusted to use instruction tokens via their respective model's chat-template.. See \autoref{tab:icl_ambiguity_cont_generating_prompt_template} for the rest of the prompt.}
\end{table*}

\begin{table*}[t]
\footnotesize
\centering
\begin{tabular}{p{0.98\textwidth}}

[...]

Question: \colorbox{apricot}{[question\_text]}

Statement: Here are some descriptions of how Ava speaks:

Answer: No, the statement does not provide any information about Ava, although it implies that following statements will contain descriptions of how they speak.

Question: \colorbox{apricot}{[question\_text]}

Statement: However, we learned that they speak brashly and with profanity, with little regard for others.

Answer: Yes, the statement describes their speaking style.

Question: \colorbox{apricot}{[question\_text]}

Statement: He carries a med-kit and a set of knives in his backpack, just in case the werewolves show up.

Answer: Yes, the statement gives us new information about what Emmanuel the Destroyer has in his bag.

Question: \colorbox{apricot}{[question\_text]}

Statement: A wide grin slowly grows across their face.

Answer: Yes, the statement describes Abul grinning.

Question: \colorbox{apricot}{[question\_text]}

Statement: Based on the story section provided, we learn that he has a mysterious book of magic.

Answer: Yes, the statement gives us the information that Dmitry has a book of magic.

Question: \colorbox{apricot}{[question\_text]}

Statement: He does not handle stress very well.

Answer: Yes, the statement describes his inability to handle stress.

Question: \colorbox{apricot}{[question\_text]}

Statement: \colorbox{apricot}{[statement]}

Answer: \\

\hline
\end{tabular}
\caption{\label{tab:icl_informative_cont_generating_prompt_template} ICL-prompt: informative (cont). See \autoref{tab:icl_informative_generating_prompt_template} for more details.}
\end{table*}

\begin{table*}[t]
\footnotesize
\centering
\begin{tabular}{p{0.98\textwidth}}

[...]

Question: \colorbox{apricot}{[question\_text]}

Statement: Based on this story snippet, they have a mysterious notebook. 

Answer: No, the statement is unambiguous in describing what Fadekemi has.

Question: \colorbox{apricot}{[question\_text]}

Statement: Daniel seems very familiar with the inner workings of the ship. 

Answer: No, the statement is unambiguous in its description of Daniel's knowledge of the ship.

Question: \colorbox{apricot}{[question\_text]}

Statement: Based on this story snippet, he greatly values the guidance of others, as evidenced by him asking for Sunita's advice. 

Answer: No, the statement is unambiguous and describes Merlin's appreciation of other people's advice.

Question: \colorbox{apricot}{[question\_text]}

Statement: They try very hard to speak with the accent of the High Aristocracy, but sometimes slip into less-idolized accent of his hometown. 

Answer: No, the statement is unambiguous and describes Abdul's accent.

Question: \colorbox{apricot}{[question\_text]}

Statement: Based on the story section provided, we can infer that he speaks with confidence and assertiveness. 

Answer: No, the statement is unambiguous in its description of Dmitry's speaking pattern.

Question: \colorbox{apricot}{[question\_text]}

Statement: He is in love with Cassian. 

Answer: No, the statement is unambiguous in who Hassan loves.

Question: \colorbox{apricot}{[question\_text]}

Statement: Furthermore, they are easily scared by the screams coming from the movie theatre. 

Answer: No, the statement is unambiguous in describing Rockefeller being scared by the screams.

Question: \colorbox{apricot}{[question\_text]}

Statement: \colorbox{apricot}{[statement]}

Answer: \\

\hline
\end{tabular}
\caption{\label{tab:icl_ambiguity_cont_generating_prompt_template} ICL-prompt: ambiguity (cont). See \autoref{tab:icl_ambiguity_generating_prompt_template} for more details.}
\end{table*}

\begin{table*}[t]
\footnotesize
\centering
\begin{tabular}{p{0.98\textwidth}}

You are a helpful and expert writing assistant. You will be given a section of a story or screenplay from the perspective of \colorbox{apricot}{[character]}. Please answer the following questions about the given statements and their relationship with the snippet provided.

Story Section:

\colorbox{apricot}{[story\_section]}

Please answer the following questions about \colorbox{apricot}{[character]} by comparing the provided statement with the story section above:

Statement: \colorbox{apricot}{[statement]}

Question: What, if any, section of the story snippet is most relevant to the given statement? Provide a brief 1-2 sentence description of this section or "N/A" if there is no relevant section.

Answer: \colorbox{apricot}{[answer]}

Question: In 1-2 sentences, compare the claim the statement makes and the section of story you highlighted in your previous answer. Are there any notable differences? Are all claims made by the statement explicitly supported? If there are no claims, write "N/A".

Answer:  \\

\hline
\end{tabular}
\caption{\label{tab:cot_generating_prompt_template} Chain-Of-Thought prompt. We use this in our Verification Module to break down the reasoning steps necessary to compare a statement about a character to a story snippet. Highlighted \colorbox{apricot}{[text]} is filled in with the appropriate information prior to generation, except for \colorbox{apricot}{[answer]} which is first generated with only the first question before being filled in here for the second. Prompts were adjusted to use instruction tokens via their respective model's chat-template.}
\end{table*}

\end{document}